\documentclass[10pt]{article}

\newif\ifpreprint
\preprinttrue

\ifpreprint
  \usepackage[preprint]{tmlr}
\else
  \usepackage{tmlr}
\fi

\usepackage{booktabs}
\usepackage{amsmath,amssymb,amsthm}
\usepackage{microtype}
\usepackage{xcolor}
\usepackage{graphicx}
\graphicspath{{figures/}}
\usepackage{subcaption}
\usepackage{multirow}
\usepackage{mathtools}
\usepackage{algorithm}
\usepackage{algpseudocode}
\usepackage{float}
\usepackage{enumitem}
\usepackage{hyperref}
\hypersetup{colorlinks=true,linkcolor=blue!60!black,citecolor=blue!60!black,urlcolor=blue!60!black}
\ifpreprint
  \hypersetup{pdftitle={Whitened Advantage Summation Policy Optimization},pdfauthor={Olaitan Olaleye, Jayson Meribe, Temidayo Adepoju}}
\fi

\newtheorem{theorem}{Theorem}
\newtheorem{proposition}{Proposition}
\newtheorem{corollary}{Corollary}
\newtheorem{assumption}{Assumption}

\newcommand{\Sig}{\boldsymbol{\Sigma}}
\newcommand{\Sighat}{\hat{\boldsymbol{\Sigma}}}
\newcommand{\Sigbar}{\bar{\boldsymbol{\Sigma}}}
\newcommand{\w}{\mathbf{w}}
\newcommand{\vv}{\mathbf{v}}
\newcommand{\uu}{\mathbf{u}}
\newcommand{\bA}{\hat{\mathbf{A}}}
\newcommand{\z}{\mathbf{z}}
\newcommand{\Imat}{\mathbf{I}}
\newcommand{\Dmat}{\mathbf{D}}
\newcommand{\Wmat}{\mathbf{W}}
\newcommand{\Umat}{\mathbf{U}}
\newcommand{\Cmat}{\mathbf{C}}
\newcommand{\Hmat}{\mathbf{H}}
\newcommand{\lmin}{\lambda_{\min}}
\newcommand{\WASPO}{\text{\mbox{WAS-PO}}}
\newcommand{\WASPOD}{\text{\mbox{WAS-PO-Diag}}}
\newcommand{\GDPO}{\text{GDPO}}
\newcommand{\GRPO}{\text{GRPO}}
\newcommand{\Ecorr}{\mathcal{E}}
\newcommand{\kw}{\kappa_{\w}}
\newcommand{\ang}{\varphi}
\newcommand{\Corr}{\operatorname{Corr}}
\newcommand{\Cov}{\operatorname{Cov}}
\newcommand{\Var}{\operatorname{Var}}
\newcommand{\cond}{\operatorname{cond}}
\newcommand{\BN}{\operatorname{BN}}

\title{Whitened Advantage Summation Policy Optimization:\\
Correlation-Aware Aggregation for\\ Multi-Reward Reinforcement Learning}

\author{\name Olaitan Olaleye \email olaitan@proofmed.ai \\
        \addr ProofMed AI
        \AND
        \name Jayson Meribe \email meribejayson@gmail.com \\
        \addr ProofMed AI and Stanford University
        \AND
        \name Temidayo Adepoju \email dayo.olaleye25@gmail.com \\
        \addr ProofMed AI}

\def\month{MM}
\def\year{YYYY}
\def\openreview{\url{https://openreview.net/forum?id=XXXX}}

\begin{document}
\maketitle

\begin{abstract}
Reinforcement learning for language models often combines several rewards by normalizing each one and adding them with fixed weights. When the rewards are correlated, the variation they share is counted once for every reward that carries it. In terms of the ZCA-decorrelated components of the reward signal, the weighted sum applies weights $\Sig^{1/2}\w$ rather than the chosen weights $\w$, where $\Sig$ is the reward correlation matrix. We study WAS-PO, which whitens the normalized advantages with $\Sig^{-1/2}$ before adding them, so that the chosen weights apply to those components. We derive the scalarized objective whose gradient it follows when normalization statistics are held fixed, show that its weights on the raw rewards can be negative, give an exact two-reward condition under which the lighter-weighted reward is reversed, and give a loose bound on the error from estimating $\Sig$ during training. On tool calling with Qwen2.5-1.5B-Instruct, WAS-PO reached a higher mean training reward than GDPO in all seven seeds with three rewards (3.8\% higher) and all three seeds with four (3.9\% higher). WAS-PO also replaces GDPO's per-prompt normalization with batch-level normalization, and we did not separate the two changes with three or four rewards. The training-reward gains did not carry over to held-out accuracy, which was slightly lower than GDPO's with three rewards (not significant) and 0.03 lower in absolute terms with four (two seeds evaluated). With two rewards, WAS-PO trailed GDPO in training reward, and a three-seed control weakly suggests that the gap comes from batch-level normalization rather than whitening. In a math setting where the two rewards were nearly identical by construction, WAS-PO drove down the lighter-weighted reward, as the sign condition predicts.
\end{abstract}

{\let\thefootnote\relax\footnotetext{Large language model assistants were used in this work for drafting and editing text, preparing and checking the bibliography, and developing code. The authors checked all derivations, results and claims and take full responsibility for the content.}}

\section{Introduction}
\label{sec:intro}

Reinforcement learning for language models now routinely optimizes several rewards at once, such as correctness, output format and length. Group Relative Policy Optimization~\citep[GRPO;][]{shao2024deepseekmath} handles this by summing the rewards and normalizing the sum within each group of completions sampled for a prompt. \citet{liu2026gdpo} showed that summing first maps distinct combinations of rewards to the same advantage, and proposed Group reward-Decoupled Normalization Policy Optimization (GDPO), which normalizes each reward within the group before taking a weighted sum.

The weights in that sum are meant to say how much each objective matters, but correlated rewards change what they do. When normalized advantages of correlated rewards are added, the variation they share enters the sum once per reward. In terms of the decorrelated components of the reward signal (Section~\ref{sec:effweight}), a weighted sum with weights $\w$ applies weights $\Sig^{1/2}\w$, where $\Sig$ is the correlation matrix of the rewards. Correlation of this kind is common in multi-reward training. In our tool-calling setup, following GDPO's conditional rewards, the format, length and structure rewards are set to zero whenever accuracy is not positive, so they rise and fall together largely by construction. At the batch-level correlations that \WASPO\ estimated in our runs, the effective weight vector of the batch-normalized sum points away from $\w$ by $9.6^\circ$, $16.0^\circ$ and $20.7^\circ$ with two, three and four rewards. Whether this is a distortion depends on how the weights are read, which we discuss in Section~\ref{sec:effweight}.

Whitening the normalized advantages with $\Sig^{-1/2}$ (ZCA whitening) before summing applies $\w$ to the decorrelated components that are closest to the individual rewards. We call the result Whitened Advantage Summation Policy Optimization (\WASPO). Whitening is not free. \WASPO\ is a scalarized policy gradient whose weights on the raw rewards, $\Dmat^{-1}\Sig^{-1/2}\w$ with $\Dmat$ the diagonal matrix of reward standard deviations, can be negative (Proposition~\ref{prop:objective}). With two rewards weighted $a>b>0$, the lighter-weighted reward receives a negative weight exactly when the correlation exceeds $2ab/(a^2+b^2)$, so the policy is trained to lower that reward, and the negative weight grows without bound as the correlation approaches one (Proposition~\ref{prop:sign}). We call this ZCA-Inversion and observe it in a math setting, where the format reward fell from 0.67 to 0.13 during training.

In our three- and four-reward tool-calling settings, \WASPO\ reached higher training reward than GDPO in all ten paired runs (seven at $K=3$, three at $K=4$). The gain did not carry over to held-out accuracy, and with two rewards \WASPO\ trailed GDPO even in training reward. \WASPO\ also replaces GDPO's per-prompt normalization with batch-level normalization, so the comparison with GDPO mixes two design changes. With two rewards, a control that keeps batch-level normalization but drops whitening (\WASPOD) scores slightly below \WASPO, which weakly suggests that normalization is the source of the two-reward gap. We have not run this control with three or four rewards, so we report the gains there as gains of \WASPO\ as a whole.

Our contributions are:
\begin{enumerate}[itemsep=0pt,topsep=2pt]
  \item an exact statement of how correlation changes the effective weights of a weighted sum of normalized advantages, with a variance decomposition (Section~\ref{sec:effweight});
  \item the objective that whitened aggregation follows, and the corresponding per-prompt objective of GDPO (Section~\ref{sec:objective});
  \item an exact two-reward condition for sign reversal, its extension to near-duplicate rewards among $K$, and an inversion index that can be logged during training (Section~\ref{sec:sign});
  \item an error bound for the estimated whitening matrix (Section~\ref{sec:theorem});
  \item experiments on tool calling with two, three and four rewards, component ablations, an observed instance of ZCA-Inversion, and runs in which a nearly singular estimate did not cause collapse (Section~\ref{sec:experiments}).
\end{enumerate}

\section{Background}
\label{sec:background}

For a prompt $x$, a policy $\pi_\theta$ samples $G$ completions $y_1,\dots,y_G$. Each completion receives a vector of $K$ rewards $\mathbf{r}_i\in\mathbb{R}^K$, and the user supplies weights $\w\in\mathbb{R}^K_{>0}$. A training batch contains $N$ completions, which we call rollouts. We write $\operatorname{mean}_g$ and $\operatorname{std}_g$ for statistics over the $G$ completions of one prompt, $\epsilon=10^{-4}$ for a constant that prevents division by zero, and $\BN(u)_i=(u_i-\operatorname{mean}(u))/(\operatorname{std}(u)+\epsilon)$ for normalization across the batch. Standard deviations are sample standard deviations, as in the trainer we use.

GRPO forms a scalar reward and normalizes it within the group:
\[
  \hat A_i^{\GRPO}=\frac{\w^\top\mathbf{r}_i-\operatorname{mean}_g(\w^\top\mathbf{r})}{\operatorname{std}_g(\w^\top\mathbf{r})+\epsilon}.
\]
GDPO normalizes each reward within the group, takes the weighted sum, and normalizes the result across the batch:
\[
  \hat A_i^{(k)}=\frac{r_i^{(k)}-\operatorname{mean}_g(r^{(k)})}{\operatorname{std}_g(r^{(k)})+\epsilon},
  \qquad
  \hat A_i^{\GDPO}=\BN\Bigl(\sum_{k=1}^K w_k\hat A^{(k)}\Bigr)_i.
\]
Within a group, each $\hat A^{(k)}$ has zero mean and close to unit variance, or is identically zero if reward $k$ is constant on that group. The vector $\bA_i=(\hat A_i^{(1)},\dots,\hat A_i^{(K)})$ keeps the correlation of the rewards. If $\Sig$ is the covariance of $\bA$ and has unit diagonal, with entries $\rho_{jk}$, the covariance of the summed advantage with the $k$-th normalized advantage is
\begin{equation}
  \Cov\bigl(\w^\top\bA,\hat A^{(k)}\bigr)=(\Sig\w)_k=w_k+\sum_{j\neq k}w_j\rho_{jk},
  \label{eq:effweight-original}
\end{equation}
so positively correlated rewards raise each other's covariance with the aggregate. Both methods update the policy with the clipped surrogate objective of PPO~\citep{schulman2017ppo}, a variant of the score-function estimator~\citep{williams1992simple}.

\section{What Weighted Summation Does to Reward Weights}
\label{sec:theory}

\subsection{Effective Weights}
\label{sec:effweight}

Let $\bA\in\mathbb{R}^K$ have zero mean and covariance $\Sig\succ0$ with unit diagonal, and let $\Sig^{\pm1/2}$ denote the symmetric positive definite square root and its inverse. Define
\begin{equation}
  \z:=\Sig^{-1/2}\bA,\qquad\text{so that}\qquad \bA=\Sig^{1/2}\z\quad\text{and}\quad\Cov(\z)=\Imat.
  \label{eq:decorr}
\end{equation}
The coordinates of $\z$ are uncorrelated, and we call them the decorrelated components of the reward signal.

Any $\Wmat$ with $\Wmat^\top\Wmat=\Sig^{-1}$ decorrelates $\bA$. The symmetric choice $\Wmat=\Sig^{-1/2}$ is zero-phase component analysis (ZCA) whitening. Because the coordinates of $\bA$ are standardized, it coincides with what \citet{kessy2018} call ZCA-cor whitening of the raw rewards, which among all whitening transforms maximizes the total correlation $\sum_k\Corr(z_k,\hat A^{(k)})$ between each component and its own reward. Principal component analysis (PCA) whitening produces components with no correspondence to individual rewards, and Cholesky whitening depends on the order in which the rewards are listed.

\begin{proposition}[Effective weights]
\label{prop:effweight}
For any $\w\in\mathbb{R}^K$,
\begin{equation}
  \w^\top\bA=\bigl(\Sig^{1/2}\w\bigr)^\top\z
  \qquad\text{and}\qquad
  \w^\top\Sig^{-1/2}\bA=\w^\top\z.
  \label{eq:effweights}
\end{equation}
Weighted summation applies weights $\Sig^{1/2}\w$ to the decorrelated components, and summation after ZCA whitening applies $\w$. The two agree for all $\w$ if and only if $\Sig=\Imat$.
\end{proposition}

The proof is in Appendix~\ref{app:proofs}. We measure the discrepancy by the angle $\ang_{\w}$ between $\Sig^{1/2}\w$ and $\w$.

Proposition~\ref{prop:effweight} does not say which weighting is appropriate, because that depends on what $\w$ is meant to weight. Let $J_k(\theta)=\mathbb{E}_{x,\,y\sim\pi_\theta}[r^{(k)}]$. If $\w$ weights the raw rewards, the target is $\sum_kw_kJ_k$ and correlation calls for no correction; summing the rewards before normalizing, as GRPO does, follows it, while summing separately normalized advantages follows $\sum_k(w_k/\sigma_k)J_k$ (Proposition~\ref{prop:objective}). If $\w$ is meant to weight the decorrelated components that track each reward, whitening is what applies those weights. We adopt the second reading, which matches the intuition that a weight such as $0.3$ on length should not be inflated because length moves with accuracy, and we examine its consequences in Section~\ref{sec:experiments}. A third reading, under which $\Cov(\text{aggregate},\hat A^{(k)})=w_k$ in~\eqref{eq:effweight-original}, would call for the regression weights $\Sig^{-1}\w$, which we do not consider further.

\begin{corollary}[Variance decomposition]
\label{cor:scale}
Let $\kw:=\w^\top\Sig\w/\|\w\|^2$. Then
\[
  \Var(\w^\top\bA)=\w^\top\Sig\w=\|\w\|^2+\sum_{j\neq k}w_jw_k\rho_{jk},\qquad \Var(\w^\top\Sig^{-1/2}\bA)=\|\w\|^2,
\]
so cross-reward terms account for a fraction $1-1/\kw$ of the variance of the summed advantage.
\end{corollary}

We computed $\ang_{\w}$ and $\kw$ from $\Sigbar$, the time average of the correlation estimate that \WASPO\ maintained during training (Table~\ref{tab:corrlog}). The angle is $9.6^\circ$, $16.0^\circ$ and $20.7^\circ$ for $K=2,3,4$, and $\kw$ is 1.52, 1.95 and 2.30, so cross-reward terms carry 34\%, 49\% and 57\% of the variance of the unwhitened sum. Because these use batch-level correlations, they describe the unwhitened sum under batch-level normalization (the \WASPOD\ control of Section~\ref{sec:method}). For GDPO, Proposition~\ref{prop:effweight} applies with $\Sig$ replaced by the pooled covariance of its group-normalized advantages, rescaled to unit diagonal; it reflects within-group correlation only, and we did not log it. Without renormalization, $\sqrt{\kw}$ would be the factor by which correlation inflates the scale of the summed advantage. GDPO and \WASPO\ both renormalize across the batch, which removes this effect, so here $\kw$ describes only the composition of the aggregate.

\subsection{The Objective Whitened Aggregation Follows}
\label{sec:objective}

Let $\vv:=\Sig^{-1/2}\w$ be the vector that whitened aggregation applies to the normalized advantages, and let $\Dmat=\operatorname{diag}(\sigma_1,\dots,\sigma_K)$ hold the reward standard deviations used for normalization.

\begin{proposition}[Effective objectives]
\label{prop:objective}
Consider the sequence-level score-function estimator without a KL penalty or length normalization. Treat all normalization statistics, the matrix $\Sig$ and the final batch scale $s>0$ (the denominator of $\BN$) as constants.
\begin{enumerate}[itemsep=0pt,topsep=2pt]
  \item[(a)] Suppose the aggregate advantage is $\tilde A_i=s^{-1}\vv^\top\bA_i$, with $\hat A_i^{(k)}=(r_i^{(k)}-\mu_k)/\sigma_k$ computed from batch statistics. Then the expected update $\mathbb{E}[\tilde A\,\nabla_\theta\log\pi_\theta(y\mid x)]$ equals $s^{-1}\nabla_\theta J_{\boldsymbol\omega}(\theta)$, where
  \begin{equation}
    J_{\boldsymbol\omega}(\theta)=\sum_{k=1}^K\omega_k\,J_k(\theta),
    \qquad
    \boldsymbol\omega=\Dmat^{-1}\Sig^{-1/2}\w.
    \label{eq:objective}
  \end{equation}
  \item[(b)] With per-prompt statistics $\mu_k(x)$, $\sigma_k(x)>0$ and $\vv=\w$, as in GDPO, the same holds with $J=\mathbb{E}_x\bigl[\sum_k\bigl(w_k/\sigma_k(x)\bigr)\,\mathbb{E}_{y\sim\pi_\theta(\cdot\mid x)}[r^{(k)}]\bigr]$.
\end{enumerate}
\end{proposition}

Under these idealizations \WASPO\ is a scalarized policy gradient with weights $\boldsymbol\omega$ on the raw rewards. The idealizations are standard but not exact. The batch statistics depend on the sampled completions (an $O(1/N)$ effect), the trainer's token-level loss differs from the sequence-level estimator, and $\boldsymbol\omega$ changes during training as the estimate of $\Sig$ and the scales $\Dmat$ change. In our runs each batch is used for a single update, so the clipping in the surrogate is inactive. Part~(b) shows that GDPO weights each reward per prompt by the inverse of its within-group standard deviation, so a reward that varies little on a prompt receives a larger weight there (at most $w_k/\epsilon$; a reward that is constant on the group contributes nothing). For binary rewards with $G=4$, the within-group standard deviation is 0.50 or 0.58 when nonzero, so this reweighting matters mainly for continuous rewards. The batch-normalized sum without whitening follows $\Dmat^{-1}\w$, whose entries are all positive. For two positively correlated rewards, and at all our measured correlation matrices, $\Sig^{-1/2}$ has negative off-diagonal entries, so $\boldsymbol\omega$ can have negative entries. A negative $\omega_k$ rewards the policy for lowering reward $k$ when the other rewards are held fixed.

\subsection{When Whitening Reverses a Reward}
\label{sec:sign}

\begin{proposition}[Sign reversal with two rewards]
\label{prop:sign}
Let $K=2$ and $\w=(a,b)$ with $a\ge b>0$, and let the correlation be $\rho\in(-1,1)$. Then $v_1>0$, and
\[
  v_2<0\quad\Longleftrightarrow\quad \rho>\rho^\star(a,b):=\frac{2ab}{a^2+b^2}.
\]
If $a>b$, then $|v_2|\to\infty$ and $v_2/v_1\to-1$ as $\rho\to1$. If $a=b$, then $\rho^\star=1$ and no reversal occurs.
\end{proposition}

The threshold depends only on the ratio of the weights. For the math setting of Section~\ref{sec:failure-nearunity}, $\w=(1.0,0.3)$ gives $\rho^\star=0.55$; for the two-reward tool-calling setting, $\w=(1.0,0.5)$ gives $\rho^\star=0.80$. As an example, at $\rho=0.98$ and $\w=(1.0,0.3)$, $\vv=(2.94,-2.01)$: an accurate but unformatted completion gains advantage relative to an accurate and formatted one. As $\rho\to1$, the aggregate approaches a positive multiple of $\hat A^{(1)}-\hat A^{(2)}$. Batch-level renormalization removes the growth in magnitude but not this direction. Corollary~\ref{cor:dup} extends the result to $K$ rewards when two of them become nearly collinear.

\begin{corollary}[Near-duplicate rewards]
\label{cor:dup}
Let $\Sig_n$ be a sequence of correlation matrices whose smallest eigenvalue $\lambda^{(n)}_1$ is simple and tends to zero, with unit eigenvector $\uu_n\to\uu$, while the other eigenvalues stay above some $\underline\lambda>0$. If $\uu^\top\w\neq0$, then $\vv_n/\|\vv_n\|\to\operatorname{sign}(\uu^\top\w)\,\uu$. In particular, if under the same hypotheses $\rho_{ij}\to1$ for one pair $i\neq j$, then $\uu=\pm(\mathbf e_i-\mathbf e_j)/\sqrt2$, where $\mathbf e_i$ is the $i$-th standard basis vector, and the whitened weights approach a multiple of $\mathbf e_i-\mathbf e_j$ whenever $w_i\neq w_j$, so the lighter-weighted of the two rewards is reversed.
\end{corollary}

For $K>2$ we summarize the sign pattern of $\vv$ with an inversion index,
\begin{equation}
  \psi(\Sig,\w):=\frac{\max\{0,-\min_kv_k\}}{\max_kv_k},
  \label{eq:psi}
\end{equation}
which is zero when all whitened weights are nonnegative and approaches one when two rewards become collinear (the pair case of Corollary~\ref{cor:dup}). The denominator is positive because $\w>0$ and $\w^\top\vv=\w^\top\Sig^{-1/2}\w>0$ force some $v_k$ to be positive. For $K=2$, $\psi<1$ because $v_1+v_2=(a+b)/\sqrt{1+\rho}>0$; for $K>2$ it can exceed one. In practice we compute $\vv$ and $\psi$ from $\vv_t=\Wmat_t\w$, which includes the eigenvalue floor of Section~\ref{sec:method}.

Table~\ref{tab:diagnostics} lists $\vv$ and $\psi$ for our settings. At $\Sigbar$, \WASPO\ places negative weights on length ($-0.15$ at $K=3$) and on length and structure ($-0.06$ and $-0.26$ at $K=4$), because both are largely predicted by accuracy and format. In all three tool-calling settings, $\psi\le0.23$ at $\Sigbar$. A negative weight on length means that \WASPO\ penalizes length beyond what accuracy and format predict; it does not by itself make the policy shorter. The estimate was sometimes much worse conditioned: its format--length entry reached 1.000 during the three- and four-reward runs (Table~\ref{tab:corrlog}). If every pair were at its largest logged value, \WASPO\ would put a large positive weight on format and a large negative weight on length at $K=3$ ($\psi=1.01$), and on structure and length at $K=4$ ($\psi=0.82$). None of the three- or four-reward runs collapsed, unlike the math runs; Section~\ref{sec:failure-nearunity} discusses why.

\begin{table}[t]
  \centering
  \caption{\textbf{Diagnostics at the estimated correlations.} $\ang_{\w}$ is the angle between $\Sig^{1/2}\w$ and $\w$ (Proposition~\ref{prop:effweight}); $\vv=\Wmat\w$ with the eigenvalue floor of Section~\ref{sec:method}; $\psi$ is the inversion index~\eqref{eq:psi}. ``Time-averaged'' rows use $\Sigbar$ (Table~\ref{tab:corrlog}). ``Largest'' rows set every pair to its largest logged value; the maxima need not occur at the same step. At $K=3,4$ these matrices are singular to numerical precision, so there the magnitudes of $\vv$ are set by the floor and only its signs and $\psi$ are meaningful. $\Delta$AUC is \WASPO\ minus GDPO.}
  \label{tab:diagnostics}
  \small\setlength{\tabcolsep}{4pt}
  \begin{tabular}{llrlrr}
    \toprule
    Setting & $\w$ & $\ang_{\w}$ & \quad$\vv$ & $\psi$ & $\Delta$AUC\\
    \midrule
    Tool, $K=2$, time-averaged & (1.0, 0.5) & $9.6^\circ$ & (1.00, 0.17) & 0.00 & $-0.014$\\
    Tool, $K=2$, largest & (1.0, 0.5) & & (1.24, $-0.14$) & 0.11 & \\
    Tool, $K=3$, time-averaged & (1.0, 0.5, 0.3) & $16.0^\circ$ & (1.05, 0.30, $-0.15$) & 0.15 & $+0.041$\\
    Tool, $K=3$, largest & (1.0, 0.5, 0.3) & & (1.22, 9.84, $-9.95$) & 1.01 & \\
    Tool, $K=4$, time-averaged & (1.0, 0.5, 0.3, 0.2) & $20.7^\circ$ & (1.10, 0.39, $-0.06$, $-0.26$) & 0.23 & $+0.049$\\
    Tool, $K=4$, largest & (1.0, 0.5, 0.3, 0.2) & & (1.32, 16.5, $-3.25$, $-13.5$) & 0.82 & \\
    Math, step 10 ($\rho\approx0.98$) & (1.0, 0.3) & & (2.94, $-2.01$) & 0.69 & $-0.228$\\
    \bottomrule
  \end{tabular}
\end{table}

\subsection{Estimation Error}
\label{sec:theorem}

In practice, $\Sig$ is estimated during training from $N$ rollouts per batch.

\begin{assumption}
\label{ass:entrywise}
An estimate $\Sighat$ of $\Sig$ from $N$ rollouts satisfies $\mathbb{E}\|\Sighat-\Sig\|_F^2\le cK^2/(N-1)$ for a constant~$c$.
\end{assumption}

Under bivariate normality, the asymptotic variance of a sample correlation is $(1-\rho^2)^2/N\le1/N$, which gives $c\approx1$. For skewed binary rewards this variance can be several times larger, and correlation among the $G$ completions of a prompt can reduce the effective sample size by up to a factor of $G$. We use $c=4$ as a working value and have not verified it for our rewards.

\begin{theorem}[Whitening estimation error]
\label{thm:error}
Let $\Sig$ be symmetric, let $\Sighat$ be a random symmetric matrix satisfying Assumption~\ref{ass:entrywise}, and suppose that almost surely all eigenvalues of $\Sig$ and $\Sighat$ are at least $\lambda_0>0$. Then
\begin{equation}
  \mathbb{E}\bigl\|(\Sighat^{-1/2}-\Sig^{-1/2})\w\bigr\|^2\le\frac{cK^2\|\w\|^2}{4\lambda_0^3(N-1)}.
  \label{eq:noisebound}
\end{equation}
\end{theorem}

The proof (Appendix~\ref{app:proof-thm1}) bounds the Fréchet derivative of $\Sig\mapsto\Sig^{-1/2}$ with the Daleckii--Krein formula~\citep[Section~V.3]{bhatia1997matrix} and integrates it along the segment from $\Sig$ to $\Sighat$. Every matrix on that segment has eigenvalues at least $\lambda_0$ because the smallest eigenvalue is a concave function of a symmetric matrix.

We compare~\eqref{eq:noisebound} with the size of the correction, $\Ecorr:=\|\Sig^{-1/2}\w-\w\|^2$, which is 0.112, 0.251 and 0.360 for $K=2,3,4$. \WASPO\ estimates $\Sig$ from all rollouts in a batch ($N=32$ in our main runs) and smooths the estimate with an exponential moving average (EMA) that gives weight $1-\beta$ to the newest batch. If successive batch estimates were independent with mean $\Sig$, the EMA started from the first batch would multiply the variance of each entry at step $t$ by
\[
  f_t=\beta^{2(t-1)}+\frac{1-\beta}{1+\beta}\bigl(1-\beta^{2(t-1)}\bigr).
\]
This factor tends to $(1-\beta)/(1+\beta)=1/199$ at $\beta=0.99$, but equals 0.38 at $t=50$, 0.14 at $t=100$ and 0.023 at $t=200$; its average over 500 steps is 0.105. Appendix~\ref{app:kstar} evaluates the bound under this idealization, with $\lambda_0$ set to the smallest eigenvalue of $\Sigbar$. At $K=3$ and $N=32$, the stationary bound is 0.26, about the size of the correction, and at $t=100$ it is 7.3. The bound is therefore too loose to rule out estimation noise comparable to the correction for much of training, and it does not apply when the estimate approaches singularity, as it did at $K=3$ and $K=4$. Its value is qualitative: the bound grows as $K^2\lmin^{-3}$, so adding near-redundant rewards (small $\lmin$) makes the whitening matrix much harder to estimate, and it falls with the effective number of rollouts behind the estimate.

\section{Method}
\label{sec:method}

Each training iteration of \WASPO\ has three stages (Algorithm~\ref{alg:waspo}), plus an optional inversion check. From here on, $\hat A^{(k)}$ denotes the batch-normalized advantage defined below.

\paragraph{Batch-level normalization.} Each reward is centered and scaled by its mean and standard deviation over all $N$ rollouts in the batch. The covariance of $\bA$ on a batch is then the batch correlation matrix $\Cmat_t$, which feeds the EMA estimate used for whitening. Batch-level normalization also changes the baseline and the scale relative to GDPO. Under GDPO a completion is compared with other completions to the same prompt, and each reward is scaled by its within-group standard deviation (Proposition~\ref{prop:objective}(b)). Under batch-level normalization, completions to easy prompts tend to receive positive advantages, much as in the global advantage normalization of REINFORCE++~\citep{hu2025reinforcepp}. The \WASPOD\ control below separates this change from whitening.

\paragraph{Correlation estimate.} The Pearson correlation matrix $\Cmat_t$ of the raw rewards is computed over the batch and replaced by $\Imat$ if any reward has zero variance in the batch. It is smoothed as $\Sighat_t=\beta\Sighat_{t-1}+(1-\beta)\Cmat_t$ with $\beta=0.99$ and $\Sighat_1=\Cmat_1$.

\paragraph{Whitening and aggregation.} With the eigendecomposition $\Sighat_t=\Umat_t\operatorname{diag}(\eta_1,\dots,\eta_K)\Umat_t^\top$, we set $\Wmat_t=\Umat_t\operatorname{diag}\bigl(\max(\eta_i,10^{-4})^{-1/2}\bigr)\Umat_t^\top$ and compute
\begin{equation}
  \hat A_i^{\WASPO}=\BN\bigl(\w^\top\Wmat_t\bA\bigr)_i.
  \label{eq:adv-canonical}
\end{equation}
The eigenvalue floor bounds $\|\Wmat_t\|_2$ by 100. It has no effect when every eigenvalue of $\Sighat_t$ exceeds $10^{-4}$, and for two rewards it removes a reversal only when $\rho>1-10^{-4}$ and $a/b<1.0142$.

\paragraph{Inversion check (optional).} Since $\vv_t=\Wmat_t\w$ is available at every step, $\psi(\Sighat_t,\w)$ can be logged at negligible cost, and $\Wmat_t$ can be replaced by $\Imat$ when needed (Section~\ref{sec:guidance}). The experiments reported here did not use this check.

\begin{algorithm}[t]
\caption{\WASPO\ advantages for one training iteration}
\label{alg:waspo}
\begin{algorithmic}[1]
  \Require rewards $\mathbf{R}\in\mathbb{R}^{N\times K}$, weights $\w$, EMA state $\Sighat$, decay $\beta$
  \State $\hat A_i^{(k)}\gets(R_{ik}-\operatorname{mean}_iR_{ik})/(\operatorname{std}_iR_{ik}+\epsilon)$
  \State $\Cmat\gets\Corr(\mathbf{R})$, or $\Cmat\gets\Imat$ if some column of $\mathbf{R}$ is constant
  \State $\Sighat\gets\Cmat$ at the first iteration, else $\Sighat\gets\beta\Sighat+(1-\beta)\Cmat$
  \State $\Wmat\gets\Umat\operatorname{diag}(\max(\eta_i,10^{-4})^{-1/2})\Umat^\top$, where $\Sighat=\Umat\operatorname{diag}(\eta)\Umat^\top$
  \State $\tilde A_i\gets\w^\top\Wmat\bA_i$; \ $\hat A_i^{\WASPO}\gets\BN(\tilde A)_i$
  \State \textit{(optional)} if the inversion check fires, recompute $\tilde A_i$ with $\Wmat=\Imat$
  \State update $\pi_\theta$ with the clipped surrogate objective using $\{\hat A_i^{\WASPO}\}$
\end{algorithmic}
\end{algorithm}

\paragraph{Controls.} \WASPOD\ zeros the off-diagonal entries of $\Sighat_t$, which sets $\Wmat_t=\Imat$ because the diagonal is one. It runs through the same code path as \WASPO, keeps batch-level normalization and drops whitening, so it differs from GDPO only in normalization and from \WASPO\ only in whitening. \WASPO-LW applies Ledoit--Wolf shrinkage~\citep{ledoit2004} to $\Sighat_t$ before inversion.

\section{Experiments}
\label{sec:experiments}

\subsection{Setup}
\label{sec:setup}

\paragraph{Training.} We train Qwen2.5-1.5B-Instruct~\citep{qwen25} for 500 steps on the single-turn subset of the Berkeley Function Calling Leaderboard v3 data~\citep[BFCL;][]{bfcl}, excluding the irrelevance category, using the copy distributed as \texttt{llamastack/bfcl\_v3}. The last 20\% of this subset, in dataset order, is held out from training. Each step samples 8 prompts with $G=4$ completions each, so $N=32$, with temperature 1.0 and at most 128 new tokens (all hyperparameters are in Table~\ref{tab:hparams}). GRPO, GDPO and \WASPO\ share one implementation, GDPO's patched version of TRL 0.18, and differ only in how advantages are computed; rewards are gathered across the eight GPUs before advantages are formed. There is no KL penalty, and each batch is used for one update. Appendix~\ref{app:hparams} gives the prompt template and other details, including a reward-wiring pitfall that reduces multi-reward training to a single reward.

\paragraph{Rewards.} The accuracy reward parses the first JSON object in a completion and compares it with the BFCL ground truth. It is 0 if no object parses, 0.1 if the function name is wrong, and otherwise $0.5$ times the fraction of ground-truth argument names present plus $0.5$ times the fraction of argument values that match. The format reward is 1 if a JSON object parses and 0 otherwise. The length reward is $\exp\{-\tfrac12(\ell/512-1)^2\}$ for a completion of $\ell$ whitespace-separated words; with completions capped at 128 tokens, it lies between 0.61 and about 0.76 and increases with length. The structure reward ($K=4$) is 1 for an object with a function name and nonempty arguments, 0.5 for a name alone, and 0 otherwise. We use GDPO's conditional rewards: every secondary reward is set to zero when the accuracy reward is not positive, and we call such rewards gated. The gated format reward is therefore exactly the indicator that accuracy is positive, and the length and structure rewards are zero whenever it is zero. This shared gate is the main source of the correlations in Table~\ref{tab:corrlog}.

\paragraph{Metrics.} Our main metric is training AUC, the mean over all $T=500$ steps of the weighted batch-mean reward,
\[
  \text{AUC}=\frac1T\sum_{t=1}^T\w^\top\bar{\mathbf r}_t,
\]
where $\bar{\mathbf r}_t$ is the batch-mean reward vector at step $t$. Its maximum is $\sum_kw_k$ (1.5, 1.8 and 2.0 for $K=2,3,4$), so AUC values are not comparable across $K$. We also report held-out accuracy: the accuracy reward of the final checkpoint under greedy decoding on the first 500 prompts of the held-out 20\%, the same prompts for every method, $K$ and seed. Seeds are paired across methods. Training comparisons use exact two-sided Wilcoxon signed-rank tests on the paired differences; with seven seeds the smallest attainable $p$-value is $2/128\approx0.016$, and with three seeds it is $0.25$. We report $\Delta$ as the mean paired difference and $d_z$ as that mean divided by the standard deviation of the paired differences. For held-out accuracy we kept per-method summaries rather than per-seed pairs, so we use unpaired two-sided Welch $t$-tests.

\subsection{Three and Four Rewards}
\label{sec:results-positive}

\begin{figure}[t]
  \centering
  \begin{subfigure}[t]{0.48\linewidth}
    \includegraphics[width=\linewidth]{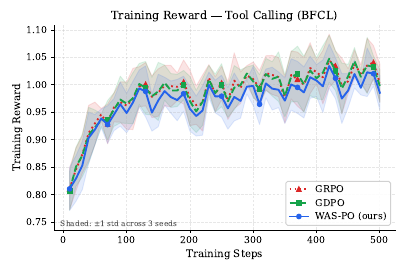}
    \caption{Two rewards.}
    \label{fig:curve-e1}
  \end{subfigure}
  \hfill
  \begin{subfigure}[t]{0.48\linewidth}
    \includegraphics[width=\linewidth]{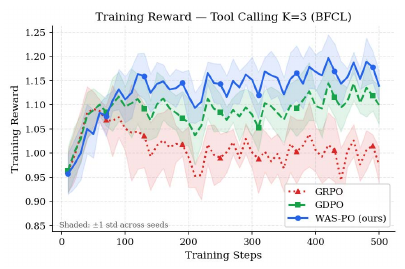}
    \caption{Three rewards.}
    \label{fig:curve-e5}
  \end{subfigure}
  \caption{\textbf{Training reward of GRPO, GDPO and \WASPO.} Curves and $\pm1$ SD bands are computed over a subset of the seeds (three at $K=2$). The per-seed AUCs in Tables~\ref{tab:e1-seeds} and~\ref{tab:e5-seeds} cover all seven seeds.}
  \label{fig:training-curves}
\end{figure}

\paragraph{Three rewards.} Figure~\ref{fig:training-curves} shows training curves for two and three rewards. The rewards are accuracy ($w_1=1.0$), format ($w_2=0.5$) and length ($w_3=0.3$). Over seven seeds, training AUC was $1.013\pm0.011$ for GRPO, $1.089\pm0.006$ for GDPO and $1.130\pm0.007$ for \WASPO\ (mean $\pm$ SD; Table~\ref{tab:e5-seeds}, Figure~\ref{fig:curve-e5}). \WASPO\ was ahead of GDPO in every seed, by $\Delta=+0.041$ ($p=0.016$, $d_z=4.6$).

Held-out accuracy was $0.397\pm0.076$ for GRPO, $0.446\pm0.012$ for GDPO and $0.438\pm0.011$ for \WASPO\ (Table~\ref{tab:heldout}). \WASPO\ was 0.008 below GDPO, which is not significant (Welch $p=0.22$) and within the standard error of a single run's mean accuracy on 500 prompts (at most about 0.022). The exact-match rate, the fraction of completions with accuracy reward 1, was 0.133 for \WASPO\ and 0.131 for GDPO, and the format rate was 0.99 for both.

\paragraph{Four rewards.} With the structure reward added ($w_4=0.2$), training AUC over three seeds was $1.116\pm0.024$ for GRPO, $1.257\pm0.007$ for GDPO and $1.306\pm0.009$ for \WASPO\ (Table~\ref{tab:e6-seeds}, Figure~\ref{fig:curve-e6}). \WASPO\ was ahead in all three seeds, by $+0.040$, $+0.050$ and $+0.058$. Three seeds cannot reach significance in a Wilcoxon test (the smallest two-sided $p$ is $0.25$), so we read this as consistent with the three-reward result rather than as independent evidence. Averaged over training, \WASPO\ was ahead on every reward: accuracy $+0.014$, format $+0.038$, length $+0.011$ and structure $+0.032$ (Table~\ref{tab:e6-per-reward}). Weighted by $\w$, about two-thirds of this gain comes from the three gated rewards; the format mean, which equals the rate of completions with positive accuracy, was 0.871 under \WASPO\ against 0.833 under GDPO. On the held-out set we recorded scores for two of the three seeds: \WASPO\ scored 0.485 and 0.487, GDPO 0.517 and 0.515, and GRPO 0.521 and 0.517. With four rewards, \WASPO\ was therefore about 0.03 below GDPO on held-out accuracy, and GRPO matched or exceeded GDPO. Relative to GDPO, the gain in training AUC was $3.8\%$ with three rewards and $3.9\%$ with four (Table~\ref{tab:kscaling}; Figure~\ref{fig:k-scaling} in Appendix~\ref{app:results}).

\begin{table}[t]
  \centering
  \caption{\textbf{Training AUC by number of rewards.} $\Delta$ is the mean paired difference, \WASPO\ minus GDPO, and Relative is $\Delta$ divided by GDPO's mean, both computed from unrounded per-seed values. $\cond(\Sigbar)$ is the condition number of the time-averaged correlation estimate. The settings differ in which rewards they contain as well as in how many, and AUC is not comparable across rows. acc, fmt, len and struct are the accuracy, format, length and structure rewards.}
  \label{tab:kscaling}
  \small\setlength{\tabcolsep}{5pt}
  \begin{tabular}{cclrrrrr}
    \toprule
    $K$ & Seeds & Rewards & GDPO & \WASPO & $\Delta$ & Relative & $\cond(\Sigbar)$\\
    \midrule
    2 & 7 & acc, fmt              & 0.985 & 0.970 & $-0.014$ & $-1.5\%$ & 4.6\\
    3 & 7 & acc, fmt, len         & 1.089 & 1.130 & $+0.041$ & $+3.8\%$ & 12.2\\
    4 & 3 & acc, fmt, len, struct & 1.257 & 1.306 & $+0.049$ & $+3.9\%$ & 16.2\\
    \bottomrule
  \end{tabular}
\end{table}

\subsection{Two Rewards and Component Ablations}
\label{sec:failure}

With accuracy and format only, training AUC over seven seeds was 0.986 for GRPO, 0.985 for GDPO and 0.970 for \WASPO, which trailed GDPO in every seed ($\Delta=-0.014$, $p=0.016$, $d_z=-4.5$; Table~\ref{tab:e1-seeds}, Figure~\ref{fig:curve-e1}). Held-out accuracy was $0.479\pm0.006$ for GRPO, $0.474\pm0.008$ for GDPO and $0.457\pm0.022$ for \WASPO\ (Welch $p=0.09$ against GDPO).

\begin{table}[t]
  \centering
  \caption{\textbf{Ablations with two rewards.} Training AUC. $\Delta$ is relative to GDPO trained with the same group size $G$ on the same seeds (seeds 42, 123 and 456 for the three-seed rows; seed 42 for the decay sweep) and is computed from unrounded values. \WASPOD\ keeps batch-level normalization and removes whitening. The decay sweep is a separate set of single-seed runs with its own GDPO baseline (0.990). Its $\beta=0.99$ run repeats the default configuration and scored 0.966, against 0.972 for seed 42 in Table~\ref{tab:e1-seeds}, so reruns at a fixed seed vary by about 0.006.}
  \label{tab:ablation}
  \small
  \begin{tabular}{lrrrr}
    \toprule
    Variant ($G=4$ unless stated) & Seeds & GDPO, same $G$ & AUC & $\Delta$\\
    \midrule
    \WASPO                   & 7 & 0.985 & 0.970 & $-0.014$\\
    \WASPO-LW                & 7 & 0.985 & 0.968 & $-0.017$\\
    \WASPOD                  & 3 & 0.986 & 0.965 & $-0.021$\\
    \WASPO, $G=8$ ($N=64$)   & 3 & 1.003 & 0.984 & $-0.019$\\
    \WASPO, $G=16$ ($N=128$) & 3 & 1.011 & 0.993 & $-0.018$\\
    \midrule
    \WASPO, $\beta=0.5$  & 1 & 0.990 & 0.967 & $-0.023$\\
    \WASPO, $\beta=0.7$  & 1 & 0.990 & 0.970 & $-0.020$\\
    \WASPO, $\beta=0.9$  & 1 & 0.990 & 0.955 & $-0.035$\\
    \WASPO, $\beta=0.95$ & 1 & 0.990 & 0.955 & $-0.035$\\
    \WASPO, $\beta=0.99$ & 1 & 0.990 & 0.966 & $-0.025$\\
    \bottomrule
  \end{tabular}
\end{table}

Changing how $\Sig$ is estimated did not close the gap (Table~\ref{tab:ablation}). EMA decays from $\beta=0.5$ to $0.99$, Ledoit--Wolf shrinkage (which lowered the largest estimated correlation from 0.87 to 0.37) and larger groups all left \WASPO\ below GDPO. At $G=8$ and at $G=16$, \WASPO\ was below GDPO in each of the three seeds (Table~\ref{tab:g-seeds}).

\WASPOD\ scored 0.965 on seeds 42, 123 and 456, while \WASPO\ scored 0.972 on the same seeds. The gap (0.007, three seeds) is about the size of the rerun variation in Table~\ref{tab:ablation}. It weakly suggests that, with two rewards, the deficit relative to GDPO comes from batch-level normalization rather than from whitening. The theory does not point to whitening as the cause either. At $\Sigbar$, $\psi=0$; the estimate started at 0.87, above $\rho^\star=0.80$, and stayed above it only for the first few dozen steps (Figure~\ref{fig:trace}, top).

In an earlier math run, the format reward was a JSON-format check applied to math answers and was therefore always zero. With a constant reward, $\Cmat_t=\Imat$ at every step and \WASPO\ reduced to \WASPOD. Its gap to GDPO was $-0.004$ over three seeds (Figure~\ref{fig:trace}, bottom; Table~\ref{tab:e2a}), so batch-level normalization does not always cost much training reward.

\subsection{ZCA-Inversion}
\label{sec:failure-nearunity}

We train the same model on the GSM8K training set~\citep{cobbe2021gsm8k}, with a system prompt asking it to end with the answer in the form \texttt{The answer is: X}. The accuracy reward is 1 if the extracted answer matches the reference and 0 otherwise; the extractor takes a boxed answer if there is one and otherwise the last number in the completion. The format reward is 1 if the completion contains \texttt{The answer is:} or a boxed answer, and 0 otherwise. Both rewards are gated as in tool calling, with $\w=(1.0,0.3)$ ($\rho^\star=0.55$); $N=32$, with at most 384 new tokens. Since the gated format reward is 1 exactly when the answer is correct and formatted, the batch correlation of the two binary rewards follows from their means $p_{\text{acc}}$ and $p_{\text{fmt}}$: $\rho=\sqrt{p_{\text{fmt}}(1-p_{\text{acc}})/(p_{\text{acc}}(1-p_{\text{fmt}}))}$ (Appendix~\ref{app:proofs}). From the means in Table~\ref{tab:e2b}, $\rho\approx0.96$ under GDPO and $0.98$ under \WASPO\ at step 10, well above $\rho^\star$. Under GDPO the means are equal at steps 50 and 500, so $\rho=1$. These values are approximate because the means are averaged over batches and seeds.

Over three seeds, training AUC was 1.050 for GRPO, 1.049 for GDPO and 0.821 for \WASPO. Under \WASPO\ the format reward fell from 0.666 at step 10 to 0.132 at step 500 while accuracy kept rising; under GDPO both rose. This matches the sign that Proposition~\ref{prop:sign} assigns to the format reward: whenever $\rho>\rho^\star$, the completions that are correct and unformatted receive the largest advantages. Because the extractor falls back to the last number, such completions are easy to produce, and the policy learned to produce them. As the format reward drops, the two rewards decouple: by the same formula, the batch correlation under \WASPO\ was about 0.41 at step 50 and 0.19 at step 500. The EMA forgets slowly (half-life 69 steps at $\beta=0.99$): starting from 0.98, the estimate cannot fall below $\rho^\star$ before step 58 unless new batches are negatively correlated, so the reversal stayed in force while the format reward fell to 0.277 by step 50.

The three- and four-reward runs show that a nearly singular estimate does not by itself cause collapse. In those runs the format--length entry of the estimate reached 1.000; this probably happened early in training, since the estimate is initialized from the first batch and the two-reward trace (Figure~\ref{fig:trace}) is highest at the start. Had every pair been at its maximum, $\psi$ would have been 0.82--1.01. Among completions that pass the gate, however, the length reward varies only between 0.61 and about 0.76, so a reversal along that direction changes the reward means little. In the math run, by contrast, the reversed reward could be switched off entirely at no cost to accuracy. Whether reward models trained on overlapping data reach correlations like those of the math rewards is untested here; ensembles of such models are known to share errors~\citep{eisenstein2023helping}.

\subsection{Why Does WAS-PO Raise the Weighted Training Reward?}
\label{sec:why}

Training AUC weights the raw rewards by $\w$. By Proposition~\ref{prop:objective}, neither method follows exactly that objective: GDPO weights reward $k$ by $w_k/\sigma_k(x)$ on each prompt, and \WASPO\ follows $\boldsymbol\omega$, which puts negative weight on length or structure at $\Sigbar$. That \WASPO\ nonetheless scores higher on $\w^\top\mathbf r$ is therefore not explained by its applying the intended weights. At $K=4$, the per-reward means show that most of the gain is in the gated rewards, through a higher rate of completions that parse and score above zero (0.871 against 0.833), which also explains why every reward moved up together. Three explanations fit. First, whitening shifts weight toward accuracy relative to $\w$ ($v_1/v_2$ is 3.6 at $K=3$ and 2.8 at $K=4$, against $w_1/w_2=2$), and every secondary reward is gated on accuracy being positive. Second, GDPO may spend gradient on noise. In a group where all four completions pass the gate, the format reward is constant and drops out, while the small length differences among them (all within 0.61--0.76) are standardized to unit variance, so by Proposition~\ref{prop:objective}(b) GDPO gives them a large weight on that prompt; batch-level normalization scales them by the much larger batch standard deviation. Third, batch-level normalization gives nonzero advantages to groups in which every completion scored the same, which GDPO does not. Comparing against \WASPOD\ and against batch-normalized summation with fixed weights $\vv$, both at $K\ge3$, would separate these explanations.

\section{Practical Guidance}
\label{sec:guidance}

Items marked \emph{observed} are supported by our experiments on one task; items marked \emph{proposed} have not been tested.

\textbf{Training reward is not held-out accuracy} (\emph{observed}). With three or four rewards, \WASPO\ reached higher training reward than GDPO in every seed but equal or lower held-out accuracy, and GRPO matched or exceeded GDPO on held-out accuracy at $K=2$ and $K=4$. In these settings the secondary rewards were close to copies of the accuracy gate. On this evidence we do not recommend \WASPO\ as a way to improve held-out accuracy; what we have shown is that it changes which weighted training objective is optimized.

\textbf{Inspect the whitened weights} (\emph{proposed}). Log $\vv_t$ and $\psi(\Sighat_t,\w)$ from the matrix that is actually inverted. Any positive $\psi$ means some reward is being reversed, but in our runs $\psi$ did not separate the collapsed run (0.69) from the worst cases assembled from per-pair maxima in runs that did not collapse (0.82--1.01), so we give no threshold. What mattered was whether the policy could move along the reversed direction cheaply. Two simple guards are to clip negative entries of $\vv_t$ to zero, or to fall back to $\Wmat_t=\Imat$ when the batch mean of a reward with $\omega_k<0$ falls over a sustained window; neither has been tested. For two rewards, Proposition~\ref{prop:sign} gives the exact boundary $\rho^\star=2ab/(a^2+b^2)$ for the correlation being whitened.

\textbf{Screen before training} (\emph{proposed}). Log rewards for a few dozen steps, estimate $\Sig$ from the batches, and compute $\ang_{\w}$, $\kw$ and $\vv$. We give no threshold for the angle; at $9.6^\circ$ \WASPO\ still differed from GDPO, mainly through normalization (Table~\ref{tab:ablation}).

\textbf{Merge near-duplicate rewards} (\emph{follows from Corollary~\ref{cor:dup}}). Two rewards that measure nearly the same thing, such as a gated format reward that is almost always earned, or two reward models trained on similar data, should be merged or left unwhitened. With unequal weights, whitening reverses the lighter-weighted one; with two equally weighted rewards it reverses neither (Proposition~\ref{prop:sign}). Our own three- and four-reward settings contain such near-duplicates (Table~\ref{tab:corrlog}), so the gains reported there come from a configuration this item advises against.

\section{Limitations}
\label{sec:limitations}

\WASPO\ changes both normalization and aggregation relative to GDPO, and the control that separates them was run only with two rewards and three seeds. The gains at $K\ge3$ are gains of \WASPO\ as a whole, and Section~\ref{sec:why} lists competing explanations that we have not tested.

The gains are in training reward only. Held-out accuracy did not improve and was lower than GDPO's with four rewards. Training AUC is measured on temperature-1 samples from the training prompts, while held-out accuracy uses greedy decoding on new prompts, so a method that sharpens the policy faster can raise the first without the second. The held-out prompts come from the same BFCL subset as the training prompts, taken in dataset order, so their category mix may differ from the training portion, and we have no evidence about transfer to other tool-calling distributions or to the official BFCL evaluation.

The tool-calling results come from one task, one 1.5B model and a small batch ($N=32$); the math runs have three seeds each. Each $K$ uses one reward set, and moving from two to three rewards adds the length reward, so the number of rewards is confounded with which rewards are used. The four-reward setting has three seeds and held-out scores for two of them. Conditional reward gating produces much of the correlation we study, and our diagnostics use batch-level correlations, not the within-group correlations that govern GDPO.

The analysis relies on a stated reading of $\w$ (Proposition~\ref{prop:effweight}), on treating normalization statistics as constants (Proposition~\ref{prop:objective}), and on Assumption~\ref{ass:entrywise} with an idealized EMA (Theorem~\ref{thm:error}). The inversion diagnostics rest on one collapsed run and two runs that did not collapse despite a nearly singular estimate.

\section{Related Work}
\label{sec:related}

\paragraph{Group-based policy optimization.} \citet{shao2024deepseekmath} introduced GRPO, which normalizes a scalar reward within a group of completions and needs no learned value function. \citet{liu2026gdpo} showed that GRPO's summed reward maps distinct reward combinations to the same advantage, and proposed GDPO. On BFCL with Qwen2.5-1.5B they report 32.81\% average accuracy for GDPO against 30.18\% for GRPO over five runs, with an evaluation protocol that differs from ours. \citet{liu2025drgrpo} analyze biases introduced by the length and standard-deviation normalization terms in GRPO, and \citet{hu2025reinforcepp} argue for normalizing advantages over the whole batch. \WASPO\ uses batch-level normalization in the spirit of the latter and adds a correction for correlation between rewards, which these methods do not consider.

\paragraph{Multiple rewards in language model alignment.} \citet{bai2022training} train with a preference model for helpfulness and harmlessness, and \citet{touvron2023llama} combine separate helpfulness and safety reward models with a threshold rule rather than a weighted sum. \citet{wu2023finegrained} combine several fine-grained reward models with fixed weights, and \citet{dai2024saferlhf} treat helpfulness and harmlessness as a constrained problem. \citet{rame2023rewarded} interpolate the weights of policies fine-tuned on different rewards to trace a Pareto front. \citet{coste2024reward} and \citet{eisenstein2023helping} combine reward models in ensembles and find that models trained on similar data share errors, and \citet{moskovitz2024confronting} find that the correlation between component reward models affects where overoptimization sets in. We instead ask how correlation changes what a fixed weight vector does.

\paragraph{Multi-task loss weighting.} Multi-task learning faces a similar problem with conflicting or redundant losses. \citet{kendall2018multi} learn task weights from homoscedastic uncertainty, \citet{sener2018multitask} cast multi-task learning as multi-objective optimization, and \citet{yu2020gradient} project away conflicting gradient components. These methods operate on losses or parameter gradients, whereas \WASPO\ changes per-sample advantages before any gradient is formed.

\paragraph{Whitening and covariance estimation.} ZCA whitening was used in early work on the independent components of natural images~\citep{bell1997independent}. \citet{kessy2018} characterize the family of whitening transforms and the optimality of ZCA, and \citet{huang2018decorrelated} use ZCA whitening inside network layers and find that PCA whitening performs worse there. Estimating covariance from few samples is a classical difficulty~\citep{ledoit2004}; in our two-reward runs, Ledoit--Wolf shrinkage did not close the gap to GDPO. \citet{amari1998natural} and \citet{kakade2001natural} precondition parameter updates with the Fisher information; \WASPO\ can loosely be viewed as preconditioning in reward space instead.

\paragraph{Reward hacking.} \citet{skalse2022} define reward hacking formally, \citet{gao2023scaling} measure how optimizing a learned reward model overshoots the reward it approximates, and \citet{pan2022} show that misspecified rewards are exploited more as agents become more capable. ZCA-Inversion differs from reward hacking in that it occurs even when every reward is correctly specified, because it arises from how the rewards are combined.

\section{Conclusion}
\label{sec:conclusion}

When normalized rewards are correlated, adding them with fixed weights does not apply those weights to the decorrelated components of the reward signal. We quantified the discrepancy, derived the objectives that GDPO and whitened aggregation follow under fixed normalization statistics, and gave an exact two-reward condition, with an extension to near-duplicate rewards, under which whitening reverses a reward. In tool calling with three or four correlated rewards, \WASPO\ reached higher training reward than GDPO in every seed, but its held-out accuracy was not higher, and with four rewards it was about 0.03 lower (two seeds). In our two-reward setting, GDPO was better. With two nearly identical rewards, \WASPO\ lowered the lighter-weighted reward, as the sign condition predicts. The open questions are how much of the training-reward gain comes from whitening rather than batch-level normalization, why it does not carry over to held-out accuracy, and whether either finding holds on other tasks and larger models.

\section*{Broader Impact Statement}

Multi-reward training is used to combine helpfulness, format and safety objectives. ZCA-Inversion is relevant to safety because whitening two highly correlated rewards with unequal weights can train the policy to lower the lighter-weighted one, which could be a safety reward. Proposition~\ref{prop:sign} gives the exact condition for two rewards in terms of the correlation being whitened, but the estimate actually inverted can differ from the population value (Table~\ref{tab:corrlog}), and negatively correlated pairs are never reversed. We ran no experiments with reward models or safety rewards. We advise against whitening a safety reward unless its weight in $\boldsymbol\omega$ is constrained to be nonnegative. We release no new datasets or models; all resources used are publicly available.

\ifpreprint
\section*{Code Availability}
Code is available at \url{https://github.com/lleye/waspo}.
\else
\section*{Code Availability}
Code will be released upon acceptance.
\fi

\bibliography{references}
\bibliographystyle{tmlr}

\appendix

\section{Proofs for Section~\ref*{sec:theory}}
\label{app:proofs}

\paragraph{Proposition~\ref*{prop:effweight}.} Since $\Sig\succ0$, its symmetric positive definite square root and the inverse of that root exist and are unique. With $\z=\Sig^{-1/2}\bA$ we have $\Cov(\z)=\Sig^{-1/2}\Sig\Sig^{-1/2}=\Imat$ and $\bA=\Sig^{1/2}\z$. Then $\w^\top\bA=\w^\top\Sig^{1/2}\z=(\Sig^{1/2}\w)^\top\z$ by symmetry, and $\w^\top\Sig^{-1/2}\bA=\w^\top\Sig^{-1/2}\Sig^{1/2}\z=\w^\top\z$. The two agree for every $\w$ if and only if $\Sig^{1/2}\w=\w$ for every $\w$, that is, $\Sig^{1/2}=\Imat$, which is equivalent to $\Sig=\Imat$. $\square$

\paragraph{Corollary~\ref*{cor:scale}.} For a fixed vector $\mathbf{m}$, $\Var(\mathbf{m}^\top\bA)=\mathbf{m}^\top\Sig\mathbf{m}$. With $\mathbf{m}=\w$ and unit diagonal this is $\|\w\|^2+\sum_{j\neq k}w_jw_k\rho_{jk}$. With $\mathbf{m}=\Sig^{-1/2}\w$ it is $\w^\top\Sig^{-1/2}\Sig\Sig^{-1/2}\w=\|\w\|^2$. $\square$

\paragraph{Proposition~\ref*{prop:objective}.} (a) With the stated constants, $\tilde A_i=s^{-1}\sum_kv_k(r_i^{(k)}-\mu_k)/\sigma_k=s^{-1}\bigl(\boldsymbol\omega^\top\mathbf{r}_i-\gamma_0\bigr)$, where $\boldsymbol\omega=\Dmat^{-1}\vv$ and $\gamma_0=\sum_kv_k\mu_k/\sigma_k$ does not depend on $y_i$. For any $q$ that does not depend on $\theta$, the score-function identity gives $\mathbb{E}_{y\sim\pi_\theta}[q(y)\nabla_\theta\log\pi_\theta(y\mid x)]=\nabla_\theta\mathbb{E}_{y\sim\pi_\theta}[q(y)]$, and $\mathbb{E}_{y\sim\pi_\theta}[\nabla_\theta\log\pi_\theta(y\mid x)]=0$. Applying both to $q=\boldsymbol\omega^\top\mathbf{r}$ and to the constant $\gamma_0$, and averaging over prompts, gives $\mathbb{E}[\tilde A\,\nabla_\theta\log\pi_\theta]=s^{-1}\nabla_\theta J_{\boldsymbol\omega}$. Setting $\Sig=\Imat$ gives $\Dmat^{-1}\w$ for the batch-normalized sum without whitening.
(b) For a fixed prompt $x$, $\tilde A_i=s^{-1}\bigl(\sum_k(w_k/\sigma_k(x))r_i^{(k)}-\gamma_0(x)\bigr)$ with $\gamma_0(x)$ independent of $y_i$, and the same argument applied conditionally on $x$, followed by averaging over prompts, gives the stated objective. Using the group mean that includes $y_i$ itself, with $\sigma_k(x)$ still held fixed, adds a factor $(1-1/G)$ common to all rewards, because the other completions are independent of $y_i$; it does not change the direction of the gradient. $\square$

\paragraph{Proposition~\ref*{prop:sign}.} For $\Sig=\begin{psmallmatrix}1&\rho\\\rho&1\end{psmallmatrix}$ the eigenvectors are $\uu_\pm=(1,\pm1)/\sqrt2$ with eigenvalues $1\pm\rho$, so $\Sig^{-1/2}=(1+\rho)^{-1/2}\uu_+\uu_+^\top+(1-\rho)^{-1/2}\uu_-\uu_-^\top$ and
\[
  v_1=\frac{a+b}{2\sqrt{1+\rho}}+\frac{a-b}{2\sqrt{1-\rho}},\qquad
  v_2=\frac{a+b}{2\sqrt{1+\rho}}-\frac{a-b}{2\sqrt{1-\rho}}.
\]
Both terms of $v_1$ are nonnegative and the first is positive, so $v_1>0$. Both terms in the expression for $v_2$ are nonnegative, so $v_2<0$ if and only if $(a-b)^2(1+\rho)>(a+b)^2(1-\rho)$, that is, $\rho\bigl[(a-b)^2+(a+b)^2\bigr]>(a+b)^2-(a-b)^2$, which simplifies to $2\rho(a^2+b^2)>4ab$, or $\rho>2ab/(a^2+b^2)$. If $a=b$, the second terms vanish and $v_1=v_2>0$. If $a>b$, as $\rho\to1$ the first terms stay bounded while the second terms grow as $(a-b)/(2\sqrt{1-\rho})$, so $|v_2|\to\infty$ and $v_2/v_1\to-1$. $\square$

\paragraph{Corollary~\ref*{cor:dup}.} Write $\vv_n=\sum_m(\lambda^{(n)}_m)^{-1/2}(\uu_{n,m}^\top\w)\,\uu_{n,m}$ in an orthonormal eigenbasis of $\Sig_n$, with $\uu_{n,1}=\uu_n$. The terms with $m\ge2$ have norm at most $\underline\lambda^{-1/2}\|\w\|$, while the first has norm $(\lambda^{(n)}_1)^{-1/2}|\uu_n^\top\w|\to\infty$ because $\uu_n^\top\w\to\uu^\top\w\neq0$. Dividing by $\|\vv_n\|$ gives the limit. For the pair case let $\mathbf d=(\mathbf e_i-\mathbf e_j)/\sqrt2$. Then $\mathbf d^\top\Sig_n\mathbf d=1-\rho_{ij}\to0$, while $\mathbf d^\top\Sig_n\mathbf d\ge\underline\lambda\sum_{m\ge2}(\uu_{n,m}^\top\mathbf d)^2$. Hence $(\uu_n^\top\mathbf d)^2\to1$, so $\uu=\pm\mathbf d$ and $\uu^\top\w=\pm(w_i-w_j)/\sqrt2$, which is nonzero when $w_i\neq w_j$. $\square$

\paragraph{Correlation of gated binary rewards.} Let $A$ and $F$ be binary with $F\le A$, as for an accuracy reward and a format reward gated on it, with means $p_{\text{acc}}$ and $p_{\text{fmt}}$. Then $\mathbb{E}[AF]=p_{\text{fmt}}$, so $\Cov(A,F)=p_{\text{fmt}}(1-p_{\text{acc}})$, and dividing by $\sqrt{p_{\text{acc}}(1-p_{\text{acc}})p_{\text{fmt}}(1-p_{\text{fmt}})}$ gives $\rho=\sqrt{p_{\text{fmt}}(1-p_{\text{acc}})/(p_{\text{acc}}(1-p_{\text{fmt}}))}$. $\square$

\section{Proof of Theorem~\ref*{thm:error}}
\label{app:proof-thm1}

\paragraph{Step 1.} Let $h(\mathbf X)=\mathbf X^{-1/2}$ on symmetric matrices with all eigenvalues at least $\lambda_0$, and write $\mathbf X=\mathbf Q\boldsymbol\Lambda\mathbf Q^\top$ with eigenvalues $\lambda_1,\dots,\lambda_K$. By the Daleckii--Krein formula~\citep[Section~V.3]{bhatia1997matrix}, the Fréchet derivative is $\mathrm{d}h(\mathbf X)[\Hmat]=\mathbf Q\bigl(\mathbf M\circ(\mathbf Q^\top\Hmat\mathbf Q)\bigr)\mathbf Q^\top$, where $\circ$ is the entrywise product and
\[
  m_{ii}=-\tfrac12\lambda_i^{-3/2},\qquad
  m_{ij}=\frac{\lambda_i^{-1/2}-\lambda_j^{-1/2}}{\lambda_i-\lambda_j}=-\frac{1}{\sqrt{\lambda_i\lambda_j}\,\bigl(\sqrt{\lambda_i}+\sqrt{\lambda_j}\bigr)}\quad(\lambda_i\neq\lambda_j),
\]
with $m_{ij}=-\frac12\lambda_i^{-3/2}$ when $\lambda_i=\lambda_j$. Since $\sqrt{\lambda_i\lambda_j}\ge\lambda_0$ and $\sqrt{\lambda_i}+\sqrt{\lambda_j}\ge2\sqrt{\lambda_0}$, every $|m_{ij}|\le1/(2\lambda_0^{3/2})$. The Frobenius norm is invariant under orthogonal conjugation and $\|\mathbf M\circ\mathbf B\|_F\le\max_{ij}|m_{ij}|\,\|\mathbf B\|_F$, so $\|\mathrm{d}h(\mathbf X)[\Hmat]\|_F\le\|\Hmat\|_F/(2\lambda_0^{3/2})$.

\paragraph{Step 2.} Let $\Hmat=\Sighat-\Sig$ and $\mathbf X_\xi=\Sig+\xi\Hmat$ for $\xi\in[0,1]$. The smallest eigenvalue is concave on symmetric matrices, so $\lmin(\mathbf X_\xi)\ge(1-\xi)\lmin(\Sig)+\xi\lmin(\Sighat)\ge\lambda_0$. Hence $h$ is continuously differentiable along the segment and $\Sighat^{-1/2}-\Sig^{-1/2}=\int_0^1\mathrm{d}h(\mathbf X_\xi)[\Hmat]\,d\xi$, which by Step 1 gives $\|\Sighat^{-1/2}-\Sig^{-1/2}\|_F\le\|\Hmat\|_F/(2\lambda_0^{3/2})$.

\paragraph{Step 3.} $\|(\Sighat^{-1/2}-\Sig^{-1/2})\w\|^2\le\|\Sighat^{-1/2}-\Sig^{-1/2}\|_F^2\|\w\|^2\le\|\Hmat\|_F^2\|\w\|^2/(4\lambda_0^3)$ almost surely. Taking expectations and applying Assumption~\ref{ass:entrywise} gives~\eqref{eq:noisebound}. $\square$

\paragraph{EMA variance factor.} Unrolling the recursion $\Sighat_t=\beta\Sighat_{t-1}+(1-\beta)\Cmat_t$ from $\Sighat_1=\Cmat_1$ gives
\[
  \Sighat_t=\beta^{t-1}\Cmat_1+(1-\beta)\sum_{j=0}^{t-2}\beta^j\Cmat_{t-j}.
\]
If the $\Cmat_t$ are independent with mean $\Sig$ and entrywise variance $\tau^2$, each entry of $\Sighat_t$ has variance $\tau^2\bigl[\beta^{2(t-1)}+(1-\beta)^2\sum_{j=0}^{t-2}\beta^{2j}\bigr]=\tau^2f_t$, with $f_t$ as in Section~\ref{sec:theorem}. Bias in the batch estimates, which is $O(1/N)$ for sample correlations, is not reduced by the EMA.

\section{Evaluating the Bound}
\label{app:kstar}

Table~\ref{tab:ksweep} evaluates~\eqref{eq:noisebound} with $c=4$ and $\lambda_0$ set to the smallest eigenvalue of $\Sigbar$, for one batch of $N$ rollouts and for the EMA under the idealization of Section~\ref{sec:theorem}, both at its stationary factor $1/199$ and at step $t=100$ ($f_{100}=0.141$). We use $N=32$ in the main runs and $N=128$ in the $G=16$ ablation. Using the time-averaged $\lmin$ is itself an idealization: the estimate became nearly singular at times in the three- and four-reward runs, and single-batch estimates can be worse, so for them Theorem~\ref{thm:error} holds only with a much smaller $\lambda_0$, or not usefully at all. At the largest two-reward estimate of 0.869, $\lmin=0.131$, and the constant $K^2\|\w\|^2/\lmin^3$ rises from 111.8 to about 2224. Step 3 of the proof is tight only when $\Sighat^{-1/2}-\Sig^{-1/2}$ is close to rank one with $\w$ in its row space.

\begin{table}[H]
  \centering\small
  \caption{\textbf{Bound~\eqref{eq:noisebound} against the size of the correction $\Ecorr$.}}
  \label{tab:ksweep}
  \setlength{\tabcolsep}{4pt}
  \begin{tabular}{crrrrrrrrr}
    \toprule
     & & & & \multicolumn{2}{c}{One batch} & \multicolumn{2}{c}{EMA, stationary} & \multicolumn{2}{c}{EMA, $t=100$}\\
    \cmidrule(lr){5-6}\cmidrule(lr){7-8}\cmidrule(lr){9-10}
    $K$ & $\lmin$ & $\Ecorr$ & $K^2\|\w\|^2/\lmin^3$ & $N=32$ & $N=128$ & $N=32$ & $N=128$ & $N=32$ & $N=128$\\
    \midrule
    2 & 0.355 & 0.112 & 111.8  & 3.61 & 0.880 & 0.018 & 0.004 & 0.508 & 0.124\\
    3 & 0.196 & 0.251 & 1601.7 & 51.7 & 12.6  & 0.260 & 0.063 & 7.29  & 1.78\\
    4 & 0.196 & 0.360 & 2932.5 & 94.6 & 23.1  & 0.475 & 0.116 & 13.3  & 3.26\\
    \bottomrule
  \end{tabular}
\end{table}

\section{Full Results}
\label{app:results}

\begin{figure}[H]
  \centering
  \includegraphics[width=0.85\linewidth]{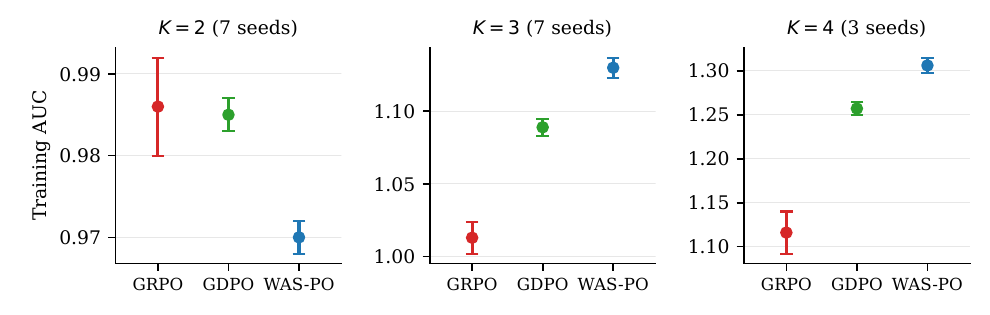}
  \caption{\textbf{Training AUC for two, three and four rewards} (mean $\pm$ SD over seeds). Each panel has its own axis because AUC is not comparable across $K$.}
  \label{fig:k-scaling}
\end{figure}

\begin{table}[H]
  \centering\small
  \caption{\textbf{Training AUC with three rewards, seven seeds.} Exact two-sided Wilcoxon $p=0.016$; $d_z=4.6$.}
  \label{tab:e5-seeds}
  \begin{tabular}{lrrrr}
    \toprule
    Seed & GRPO & GDPO & \WASPO & $\Delta$\\\midrule
    42   & 1.0303 & 1.0958 & 1.1252 & $+0.0294$\\
    123  & 0.9995 & 1.0862 & 1.1386 & $+0.0524$\\
    456  & 1.0028 & 1.0903 & 1.1213 & $+0.0310$\\
    789  & 1.0188 & 1.0947 & 1.1391 & $+0.0444$\\
    1011 & 1.0059 & 1.0798 & 1.1303 & $+0.0505$\\
    2024 & 1.0110 & 1.0863 & 1.1246 & $+0.0383$\\
    9999 & 1.0225 & 1.0879 & 1.1315 & $+0.0436$\\
    \midrule
    Mean $\pm$ SD & $1.013\pm0.011$ & $1.089\pm0.006$ & $1.130\pm0.007$ & $+0.041$\\
    \bottomrule
  \end{tabular}
\end{table}

\begin{table}[H]
  \centering\small
  \caption{\textbf{Training AUC with four rewards, three seeds.} $d_z=5.7$; the smallest attainable two-sided Wilcoxon $p$ with three seeds is 0.25.}
  \label{tab:e6-seeds}
  \begin{tabular}{lrrrr}
    \toprule
    Seed & GRPO & GDPO & \WASPO & $\Delta$\\\midrule
    42   & 1.1049 & 1.2633 & 1.3036 & $+0.0403$\\
    123  & 1.0995 & 1.2488 & 1.2987 & $+0.0499$\\
    456  & 1.1439 & 1.2579 & 1.3154 & $+0.0575$\\
    \midrule
    Mean $\pm$ SD & $1.116\pm0.024$ & $1.257\pm0.007$ & $1.306\pm0.009$ & $+0.049$\\
    \bottomrule
  \end{tabular}
\end{table}

\begin{table}[H]
  \centering\small
  \caption{\textbf{Training AUC with two rewards, seven seeds.} Exact two-sided Wilcoxon $p=0.016$; $d_z=-4.5$.}
  \label{tab:e1-seeds}
  \begin{tabular}{lrrrr}
    \toprule
    Seed & GRPO & GDPO & \WASPO & $\Delta$\\\midrule
    42   & 0.9903 & 0.9870 & 0.9721 & $-0.0149$\\
    123  & 0.9759 & 0.9826 & 0.9722 & $-0.0104$\\
    456  & 0.9918 & 0.9873 & 0.9713 & $-0.0160$\\
    789  & 0.9898 & 0.9833 & 0.9727 & $-0.0106$\\
    1011 & 0.9825 & 0.9870 & 0.9691 & $-0.0179$\\
    2024 & 0.9825 & 0.9814 & 0.9683 & $-0.0131$\\
    9999 & 0.9908 & 0.9849 & 0.9668 & $-0.0181$\\
    \midrule
    Mean $\pm$ SD & $0.986\pm0.006$ & $0.985\pm0.002$ & $0.970\pm0.002$ & $-0.014$\\
    \bottomrule
  \end{tabular}
\end{table}

\begin{table}[H]
  \centering\small
  \caption{\textbf{Training AUC at larger group sizes, two rewards, three seeds.} Eight prompts per step, so $N=64$ at $G=8$ and $N=128$ at $G=16$.}
  \label{tab:g-seeds}
  \begin{tabular}{llrrrr}
    \toprule
    $G$ & Seed & GRPO & GDPO & \WASPO & $\Delta$\\\midrule
    8  & 42  & 1.0021 & 1.0061 & 0.9849 & $-0.0212$\\
    8  & 123 & 0.9964 & 0.9976 & 0.9837 & $-0.0139$\\
    8  & 456 & 1.0020 & 1.0050 & 0.9830 & $-0.0220$\\
    \midrule
    16 & 42  & 1.0125 & 1.0111 & 0.9938 & $-0.0173$\\
    16 & 123 & 1.0084 & 1.0064 & 0.9905 & $-0.0159$\\
    16 & 456 & 1.0169 & 1.0153 & 0.9960 & $-0.0193$\\
    \bottomrule
  \end{tabular}
\end{table}

\begin{table}[H]
  \centering\small
  \caption{\textbf{Held-out accuracy} (accuracy reward of the final checkpoint under greedy decoding on 500 held-out prompts; mean $\pm$ SD over seeds). $p$ is from an unpaired two-sided Welch $t$-test computed from these summaries. At $K=4$, scores were recorded for two of three seeds.}
  \label{tab:heldout}
  \begin{tabular}{lcccrr}
    \toprule
    Setting & GRPO & GDPO & \WASPO & $\Delta$ vs.\ GDPO & $p$\\\midrule
    $K=2$ (seven seeds) & $0.479\pm0.006$ & $0.474\pm0.008$ & $0.457\pm0.022$ & $-0.017$ & 0.09\\
    $K=3$ (seven seeds) & $0.397\pm0.076$ & $0.446\pm0.012$ & $0.438\pm0.011$ & $-0.008$ & 0.22\\
    $K=4$ (two seeds) & 0.521, 0.517 & 0.517, 0.515 & 0.485, 0.487 & $-0.030$ & ---\\
    \bottomrule
  \end{tabular}
\end{table}

\begin{table}[H]
  \centering\small
  \caption{\textbf{Training AUC in the earlier math run with a constant format reward,} where \WASPO\ reduces to \WASPOD\ (Section~\ref{sec:failure}). This is not the run of Section~\ref{sec:failure-nearunity}.}
  \label{tab:e2a}
  \begin{tabular}{lrrrr}
    \toprule
    Seed & GRPO & GDPO & \WASPO & $\Delta$\\\midrule
    42  & 0.8077 & 0.8041 & 0.8085 & $+0.0044$\\
    123 & 0.7988 & 0.8120 & 0.7943 & $-0.0177$\\
    456 & 0.8075 & 0.7994 & 0.7995 & $+0.0001$\\
    \midrule
    Mean & 0.805 & 0.805 & 0.801 & $-0.004$\\
    \bottomrule
  \end{tabular}
\end{table}

\begin{table}[H]
  \centering\small
  \caption{\textbf{Per-reward batch-mean training reward in the math setting} (three seeds).}
  \label{tab:e2b}
  \begin{tabular}{lrrrr}
    \toprule
    Step & \WASPO\ acc & \WASPO\ fmt & GDPO acc & GDPO fmt\\\midrule
    10  & 0.675 & 0.666 & 0.691 & 0.675\\
    50  & 0.697 & 0.277 & 0.775 & 0.775\\
    500 & 0.806 & 0.132 & 0.863 & 0.863\\
    \bottomrule
  \end{tabular}
\end{table}

\begin{table}[H]
  \centering\small
  \caption{\textbf{Per-reward training reward with four rewards,} averaged over 500 steps (three seeds). Per-reward values were not logged for GRPO. The weighted sums of these means (1.262 and 1.305) agree with the AUCs in Table~\ref{tab:e6-seeds} to within 0.006.}
  \label{tab:e6-per-reward}
  \begin{tabular}{lrrrr}
    \toprule
    Method & Accuracy & Format & Length & Structure\\\midrule
    GDPO   & 0.521 & 0.833 & 0.528 & 0.831\\
    \WASPO & 0.535 & 0.871 & 0.539 & 0.863\\
    \midrule
    $\Delta$ & $+0.014$ & $+0.038$ & $+0.011$ & $+0.032$\\
    \bottomrule
  \end{tabular}
\end{table}

\begin{table}[H]
  \centering\small
  \caption{\textbf{The correlation estimate $\Sighat_t$ maintained by \WASPO} (seed 42 for $K=2,3$; three seeds for $K=4$). Mean and range are over the 500 training steps; $\Sigbar$ is the matrix of means, and its eigenvalues, $\cond(\Sigbar)$ and $\Ecorr$ are listed.}
  \label{tab:corrlog}
  \begin{tabular}{clrrrr}
    \toprule
    $K$ & Pair & Mean $\rho$ & Range & $\cond(\Sigbar)$ & $\Ecorr$\\\midrule
    2 & acc--fmt & 0.645 & [0.530, 0.869] & 4.6 & 0.112\\
    \multicolumn{2}{l}{\emph{eigenvalues}} & \multicolumn{3}{l}{$[0.355,\ 1.645]$} &\\\midrule
    3 & acc--fmt & 0.643 & [0.545, 0.828] & \multirow{3}{*}{12.2} & \multirow{3}{*}{0.251}\\
      & acc--len & 0.642 & [0.544, 0.827] &&\\
      & fmt--len & 0.804 & [0.698, 1.000] &&\\
    \multicolumn{2}{l}{\emph{eigenvalues}} & \multicolumn{3}{l}{$[0.196,\ 0.408,\ 2.396]$} &\\\midrule
    4 & acc--fmt    & 0.652 & [0.561, 0.828] & \multirow{6}{*}{16.2} & \multirow{6}{*}{0.360}\\
      & acc--len    & 0.651 & [0.560, 0.827] &&\\
      & acc--struct & 0.643 & [0.551, 0.828] &&\\
      & fmt--len    & 0.804 & [0.700, 1.000] &&\\
      & fmt--struct & 0.796 & [0.693, 1.000] &&\\
      & len--struct & 0.796 & [0.694, 1.000] &&\\
    \multicolumn{2}{l}{\emph{eigenvalues}} & \multicolumn{3}{l}{$[0.196,\ 0.207,\ 0.420,\ 3.177]$} &\\
    \bottomrule
  \end{tabular}
\end{table}

\begin{figure}[H]
  \centering
  \includegraphics[width=0.6\linewidth]{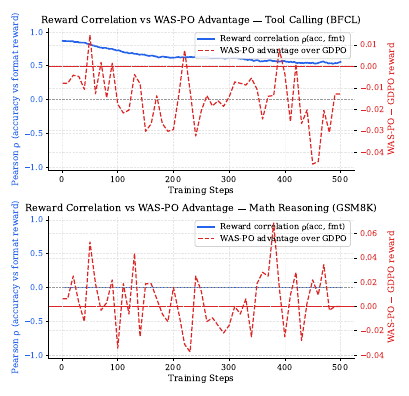}
  \caption{\textbf{Correlation estimate and reward difference} (seed 42, logged every 10 steps). Blue: the EMA estimate of $\rho(\text{acc},\text{fmt})$. Red: training reward of \WASPO\ minus GDPO, labeled ``advantage'' in the legend but meaning a difference in reward. Top: tool calling with two rewards; the two series have correlation 0.44, but both trend over training, so we do not interpret this. Bottom: the earlier math run with a constant format reward, where the estimate is the identity and the difference fluctuates around zero.}
  \label{fig:trace}
\end{figure}

\begin{figure}[H]
  \centering
  \includegraphics[width=0.6\linewidth]{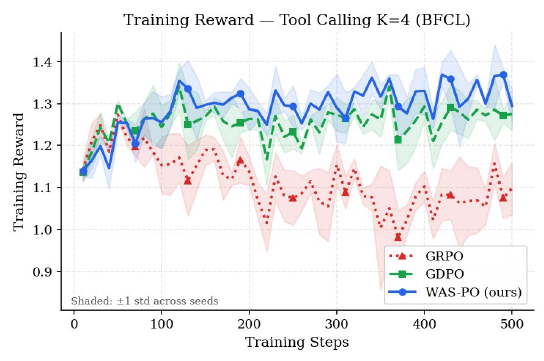}
  \caption{\textbf{Training reward with four rewards} (three seeds, mean $\pm$ SD).}
  \label{fig:curve-e6}
\end{figure}

\section{Hyperparameters and Implementation Details}
\label{app:hparams}

\begin{table}[H]
  \centering\small
  \caption{\textbf{Training hyperparameters.} Settings not listed are the defaults of TRL 0.18 and Transformers.}
  \label{tab:hparams}
  \begin{tabular}{lcccc}
    \toprule
    & Tool, $K=2$ & Tool, $K=3$ & Tool, $K=4$ & Math\\\midrule
    Model & \multicolumn{4}{c}{Qwen2.5-1.5B-Instruct}\\
    Data & \multicolumn{3}{c}{BFCL v3, single-turn, irrelevance excluded; first 80\%} & GSM8K train\\
    Group size $G$ & 4 & 4 & 4 & 4\\
    Prompts per step & 8 & 8 & 8 & 8\\
    Rollouts per step $N$ & 32 & 32 & 32 & 32\\
    Max new tokens & 128 & 128 & 128 & 384\\
    Max prompt length & \multicolumn{4}{c}{512 tokens}\\
    Training steps & 500 & 500 & 500 & 500\\
    Seeds & 7 & 7 & 3 & 3\\
    Reward weights & (1.0, 0.5) & (1.0, 0.5, 0.3) & (1.0, 0.5, 0.3, 0.2) & (1.0, 0.3)\\
    Conditional (gated) rewards & \multicolumn{4}{c}{yes}\\
    EMA decay $\beta$ & \multicolumn{4}{c}{0.99}\\
    Eigenvalue floor; $\epsilon$ & \multicolumn{4}{c}{$10^{-4}$; $10^{-4}$}\\
    KL coefficient & \multicolumn{4}{c}{0 (no reference model)}\\
    Learning rate & \multicolumn{4}{c}{$1\times10^{-6}$}\\
    Optimizer & \multicolumn{4}{c}{AdamW, linear decay, no warmup}\\
    Updates per batch & \multicolumn{4}{c}{1}\\
    Sampling & \multicolumn{4}{c}{temperature 1.0, top-$p$ 1.0}\\
    Held-out decoding & \multicolumn{4}{c}{greedy}\\
    Precision & \multicolumn{4}{c}{bfloat16}\\
    Framework & \multicolumn{4}{c}{TRL 0.18 with the GDPO patch}\\
    Hardware & \multicolumn{4}{c}{$8\times$H100}\\
    \bottomrule
  \end{tabular}
\end{table}

\paragraph{Prompts.} Tool-calling prompts list the names and descriptions (truncated to 80 characters) of the first three available functions, then the user request (truncated to 300 characters), then the instruction to respond only with a JSON function call, followed by the example \texttt{\{"name": "function\_name", "arguments": \{"param": "value"\}\}}. The held-out evaluation uses the same content with the system message ``You are a helpful assistant.'' Ground-truth argument values in BFCL can list several acceptable values; the accuracy reward compares against the first, as a case-insensitive string match. The math system prompt asks the model to end its response with \texttt{The answer is: X}, where X is the numeric answer only.

\paragraph{Compute.} The reported runs used about 120 GPU-hours, and preliminary and failed runs about 40 more. The $G=8$ and $G=16$ ablations also use eight prompts per step.

\paragraph{Reward wiring.} A natural way to add several rewards to a GRPO trainer is to wrap all reward functions in one function that returns their weighted sum. The trainer then sees a single reward, and GRPO, GDPO and \WASPO\ all run the same single-reward path, so no comparison between them is meaningful. All results here use an implementation that passes each reward function to the trainer separately, so that the reward tensor has one column per reward. We recommend checking that the reward tensor has $K$ columns before comparing multi-reward methods.

\end{document}